\documentclass{article}

\usepackage[accepted]{icml2019}

\usepackage{microtype}
\usepackage{subcaption}

\usepackage{booktabs} 
\usepackage{amsmath}
\usepackage{graphicx}
\usepackage{multirow}
\usepackage[utf8]{inputenc} 
\usepackage[T1]{fontenc}    
\usepackage{url}            
\usepackage{amsfonts}       
\usepackage{nicefrac}       
\usepackage[noend]{algpseudocode}
\usepackage[ruled,vlined,linesnumbered]{algorithm2e}
\let\oldnl\nl
\newcommand{\nonl}{\renewcommand{\nl}{\let\nl\oldnl}}
\usepackage{url}
\usepackage{bbm}
\makeatletter
\def\BState{\State\hskip-\ALG@thistlm}
\makeatother

\usepackage{hyperref}
\icmltitlerunning{Area Attention}
\begin{document}
\twocolumn[
\icmltitle{Area Attention}
\begin{icmlauthorlist}
\icmlauthor{Yang Li}{goo}
\icmlauthor{Lukasz Kaiser}{goo}
\icmlauthor{Samy Bengio}{goo}
\icmlauthor{Si Si}{goo}
\end{icmlauthorlist}
\icmlaffiliation{goo}{Google Research, Mountain View, CA, USA}
\icmlcorrespondingauthor{Yang Li}{liyang@google.com}
\icmlkeywords{Machine Learning, ICML}
\vskip 0.3in
]
\printAffiliationsAndNotice

\begin{abstract}
  Existing attention mechanisms are trained to attend to individual items in a collection (the memory) with a predefined, fixed granularity, e.g., a word token or an image grid. We propose area attention: a way to attend to areas in the memory, where each area contains a group of items that are structurally adjacent, e.g., spatially for a 2D memory such as images, or temporally for a 1D memory such as natural language sentences. Importantly, the shape and the size of an area are dynamically determined via learning, which enables a model to attend to information with varying granularity. Area attention can easily work with existing model architectures such as multi-head attention for simultaneously attending to multiple areas in the memory. We evaluate area attention on two tasks: neural machine translation (both character and token-level) and image captioning, and improve upon strong (state-of-the-art) baselines in all the cases. These improvements are obtainable with a basic form of area attention that is parameter free.
\end{abstract}

\section{Introduction}

Attentional mechanisms have significantly boosted the accuracy on a variety of deep learning tasks \citep{DBLP:journals/corr/BahdanauCB14, DBLP:journals/corr/LuongPM15,DBLP:journals/corr/XuBKCCSZB15}. They allow the model to selectively focus on specific pieces of information, which can be a word in a sentence for neural machine translation \citep{DBLP:journals/corr/BahdanauCB14, DBLP:journals/corr/LuongPM15} or a region of pixels in image captioning \citep{DBLP:journals/corr/XuBKCCSZB15,DBLP:conf/acl/SoricutDSG18}. 

An attentional mechanism typically follows a \textit{memory-query} paradigm, where the memory $M$ contains a collection of items of information from a source modality such as the embeddings of an image \citep{DBLP:journals/corr/XuBKCCSZB15} or the hidden states of encoding an input sentence \citep{DBLP:journals/corr/BahdanauCB14, DBLP:journals/corr/LuongPM15}, and the query $q$ comes from a target modality such as the hidden state of a decoder model. In recent architectures such as Transformer \citep{DBLP:journals/corr/VaswaniSPUJGKP17}, self-attention involves queries and memory from the same modality for either encoder or decoder. Each item in the memory has a key-value pair, $(k_i,v_i)$, where the key is used to compute the probability $a_i$ regarding how well the query matches the item (see Equation \ref{alignment}).
\begin{equation}
\label{alignment}
    a_i=\dfrac{\exp(f_{att}(q,k_i))}{\sum_{j=1}^{|M|}\exp(f_{att}(q,k_j))}
\end{equation}
The typical choices for $f_{att}$ include dot products $qk_i$ \citep{DBLP:journals/corr/LuongPM15} and a multilayer perceptron \citep{DBLP:journals/corr/BahdanauCB14}. The output $O_q^M$ from querying the memory $M$ with $q$ is then calculated as the sum of all the values in the memory weighted by their probabilities (see Equation \ref{memory_output}), which can be fed to other parts of the model for further calculation. During training, the model learns to attend to specific pieces of information given a query. For example, it can associate a word in the target sentence with a word in the source sentence for translation tasks.

\begin{equation}
\label{memory_output}
    O_q^M =\sum_{i=1}^{|M|}a_iv_i
\end{equation}

Attention mechanisms are typically designed to focus on individual items in the entire memory, where each item defines the granularity of what the model can attend to. For example, it can be a character for a character-level translation model, a word for a word-level model, a grid cell for an image-based model or a hidden state in a latent space. Such a construction of attention granularity is predetermined rather than learned. While this kind of item-based attention has been helpful for many tasks, it can be fundamentally limited for modeling complex attention distribution that might be involved in a task. 

In this paper, we propose \textit{area attention}, as a general mechanism for the model to attend to a group of items in the memory that are structurally adjacent. In area attention, each unit for attention calculation is an area that can contain one or more than one item. Because each of these areas can aggregate a varying number of items, the granularity of attention is learned from the data rather than predetermined. Note that area attention subsumes item-based attention because when an area contains a single item, it is equivalent to regular attention mechanisms. Area attention can be used along multi-head attention \citep{DBLP:journals/corr/VaswaniSPUJGKP17}. With each head using area attention, multi-head area attention allows the model to attend to multiple areas in the memory. As we show in the experiments, the combination of both achieved the best results.

Extensive experiments with area attention indicate that area attention outperforms regular attention on a number of recent models for two popular tasks: machine translation (both token and character-level translation on WMT'14 EN-DE and EN-FR), and image captioning (trained on COCO and tested for both in-domain with COCO40 and out-of-domain captioning with Flickr 1K). These models involve several distinct architectures, such as the canonical LSTM seq2seq with attention \citep{DBLP:journals/corr/LuongPM15} and the encoder-decoder Transformer \citep{DBLP:journals/corr/VaswaniSPUJGKP17, DBLP:conf/acl/SoricutDSG18}.

\section{Related Work}
The issue of item-grouping such as ranges or segments of a sentence, beyond individual tokens, has been investigated for problems such as dependency parsing or constituency parsing in natural language processing. Recent works \citep{DBLP:conf/acl/WangC16a, DBLP:journals/corr/SternAK17, DBLP:journals/corr/abs-1805-01052} represent a sentence segment by subtracting the encoding of the first token from that of the last token in the segment, assuming the encoder captures contextual dependency of tokens. The popular choices of the encoder are LSTM \citep{DBLP:conf/acl/WangC16a, DBLP:journals/corr/SternAK17} or Transformer \citep{DBLP:journals/corr/abs-1805-01052}. In contrast, the representation of an area (or a segment) in area attention, for its basic form, is defined as the mean of all the vectors in the segment where each vector does not need to carry contextual dependency. We calculate the mean of each area of vectors using subtraction operation over a summed area table \citep{Viola01rapidobject} that is fundamentally different from the subtraction applied in these previous works.

Lee et al. proposed a rich representation for a segment in coreference resolution tasks \citep{DBLP:journals/corr/LeeHLZ17}, where each span (segment) in a document is represented as a concatenation of the encodings of the first and last words in the span, the size of the span and an attention-weighted sum of the word embeddings within the span. Again, this approach operates on encodings that have already captured contextual dependency between tokens, while area attention we propose does not require each item to carry contextual or dependency information. In addition, the concept of range, segment or span in all the above works is proposed in a specific context and addresses a unique language-related task, rather than aiming for improving general attentional mechanisms that can be applied to any problems.

Instead of using softmax as attention activation function, sigmoid has been used to allow multiple items to be attended \citep{DBLP:journals/corr/ShenL16, rei-sogaard-2018-zero}. An important distinction is that using sigmoid activation alone does not enforce the constraint for attended items to be structurally adjacent while area attention does.

Previous works have proposed several methods for capturing structures in attention calculation. For example, Kim et al. used a conditional random field to directly model the dependency between items, which allows multiple "cliques" of items to be attended to at the same time \citep{DBLP:journals/corr/KimDHR17}. Niculae and Blondel approached the problem, from a different angle, by using regularizers to encourage attention to be placed onto contiguous segments \citep{NIPS2017_6926}. In image captioning tasks, previous work showed that it is beneficial to attend to semantic regions or concepts on an image \citep{DBLP:journals/corr/PedersoliLSV16, learning-multi-attention-convolutional-neural-network-fine-grained-image-recognition,DBLP:journals/corr/AndersonHBTJGZ17,DBLP:journals/corr/abs-1803-09845, DBLP:journals/corr/YouJWFL16}. They often train a dedicated sub-network such as Fast R-CNN \citep{DBLP:journals/corr/Girshick15} to extract region or object proposals.

Compared to these previous works, area attention we propose here does not require to train a special network or sub-network, or use an additional loss (regularizer) to capture structures, and can be entirely parameter free. It allows a model to attend to information at a varying granularity, which can be at the input layer where each item might lack contextual information, or in the latent space. While region proposal-based methods can probably extract better-quality regions as they are often pre-trained with labeled image regions, area attention is more lightweight and generally applicable. It is easy to apply area attention to existing single or multi-head attention mechanisms. By enhancing Transformer, an attention-based architecture, \citep{DBLP:journals/corr/VaswaniSPUJGKP17} with area attention, we achieved state-of-art results on a number of tasks.

\section{Area-Based Attention Mechanisms}
\begin{figure}[ht]
  \centering
  \includegraphics[width=0.80\linewidth]{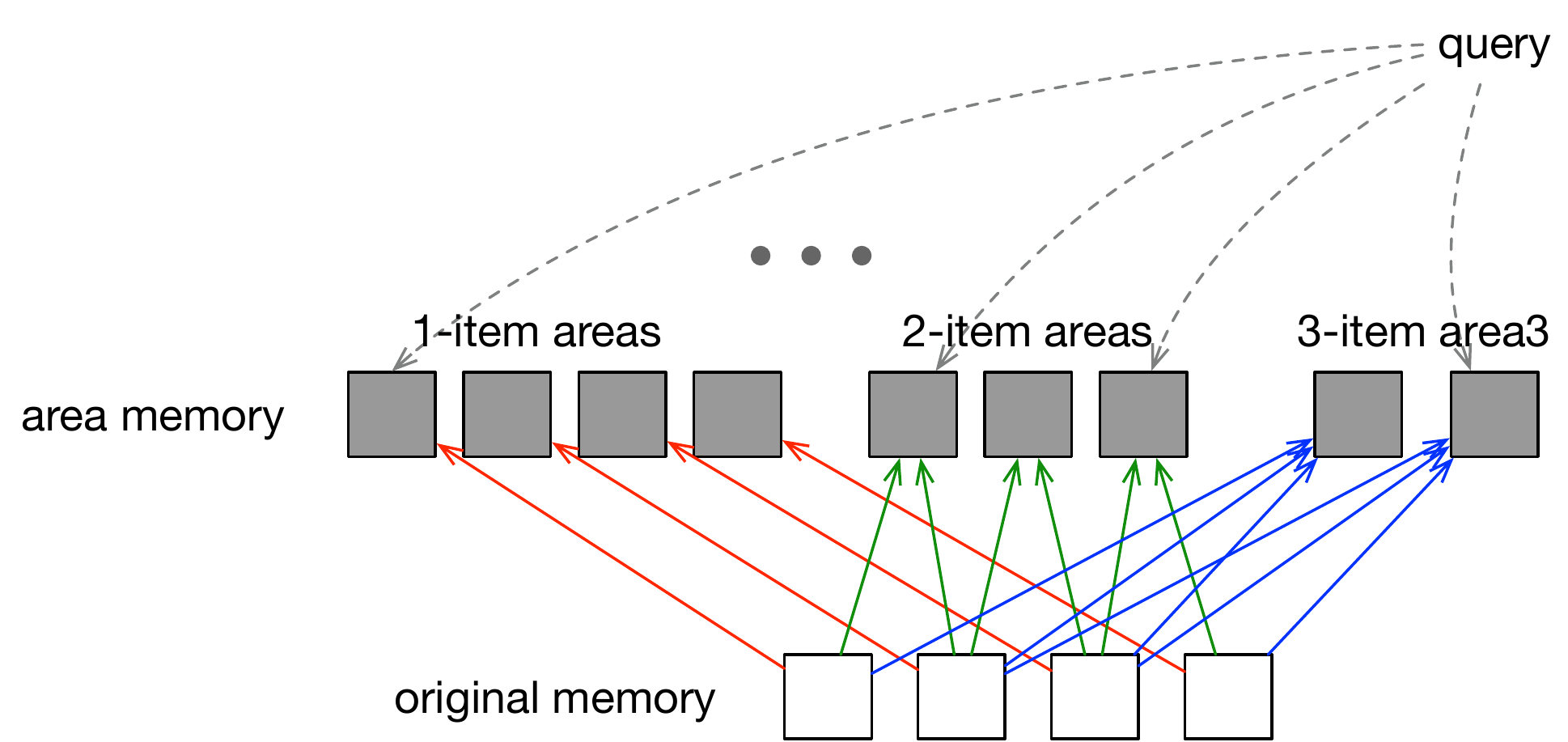}
  \caption{\label{area-1d}An illustration of area attention for the 1-dimensional case. In this example, the memory is a 4-item sequence and the maximum size of an area allowed is 3.}
\end{figure}

An area is a group of structurally adjacent items in the memory. When the memory consists of a sequence of items, a 1-dimensional structure, an area is a range of items that are sequentially (or temporally) adjacent and the number of items in the area can be one or multiple. Many language-related tasks are categorized in the 1-dimensional case, e.g., machine translation or sequence prediction tasks. In Figure \ref{area-1d}, the original memory is a 4-item sequence. By combining the adjacent items in the sequence, we form area memory where each item is a combination of multiple adjacent items in the original memory. We can limit the maximum area size to consider for a task, e.g., 3 in Figure \ref{area-1d}. 

\begin{figure}[ht]
  \centering
  \includegraphics[width=0.80\linewidth]{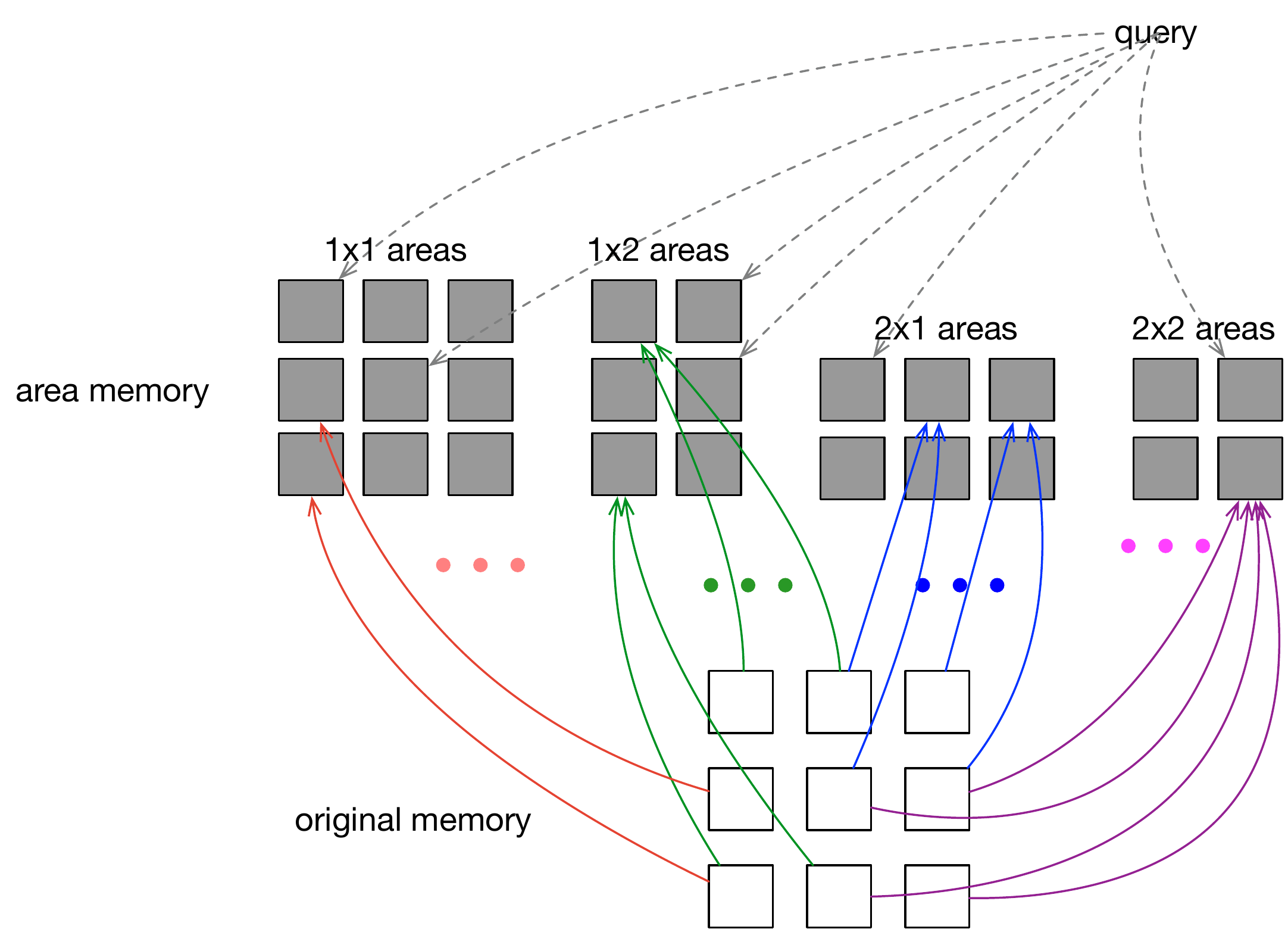}
  \caption{\label{area-2d}An illustration of area attention for the 2-dimensional case. In this example, the memory is a 3x3 grid and the dimension allowed for an area is 2x2.}
\end{figure}

When the memory contains a grid of items, a 2-dimensional structure, an area can be any rectangular region in the grid (see Figure \ref{area-2d}). This resembles many image-related tasks, e.g., image captioning. Again, we can limit the maximum size allowed for an area. For a 2-dimensional area, we can set the maximum height and width for each area. In this example, the original memory is a 3x3 grid of items and the maximum height and width allowed for each area is 2.

As we can see, many areas can be generated by combining adjacent items. For the 1-dimensional case, the number of areas that can be generated is $|R|=(L-S)S+(S+1)S/2$ where $S$ is the maximum size of an area and $L$ is the length of the sequence. For the 2-dimensional case, there are an quadratic number of areas can be generated from the original memory: $|R|=|R_{v}||R_{h}|$. $|R_{v}|=(L_{v}-H)H+(H+1)H/2$ and $|R_{h}|=(L_{h}-W)W+(W+1)W/2$ where $L_{v}$ and $L_{h}$ are the height and width of the memory grid and $H$ and $W$ are the maximum height and width allowed for a rectangular area. 

To be able to attend to each area, we need to define the key and value for each area that contains one or multiple items in the original memory. As the first step to explore area attention, we define the key of an area, $\mu_{i}$, simply as the mean vector of the key of each item in the area.

\begin{equation}
\label{area_key_mean}
    \mu_{i}=\frac{1}{|r_i|}\sum_{j=1}^{|r_i|}{k_{i,j}}
\end{equation}

where $|r_i|$ is the size of the area $r_i$. For the value of an area, we define it as the the sum of all value vectors in the area.

\begin{equation}
\label{area_value}
    v_i^{r_i}=\sum_{j=1}^{|r_i|}{v_{i,j}}
\end{equation}

With the keys and values defined, we can use the standard way for calculating attention as discussed in Equation \ref{alignment} and Equation \ref{memory_output}. Note that this basic form of area attention (Eq.\ref{area_key_mean} and Eq.\ref{area_value}) is parameter-free---it does not introduce any parameters to be learned. Essentially, Equation \ref{area_key_mean} and \ref{area_value} use average and sum pooling over an area of vectors. It is possible to use other pooling methods to compute the key and value vector for each area such as max pooling, which we will discuss later.

\subsection{Combining Area Features}

Alternatively, we can derive a richer representation of each area by using features other than the mean of the key vectors of the area. For example, we can consider the standard deviation of the key vectors within each area.

\begin{equation}
\label{area_var}
    \sigma_{i} =\sqrt{\dfrac{1}{|r_i|}\sum_{l=1}^{|r_i|}{(k_{i,l}-\mu_{i})^2}}
\end{equation}



We can also consider the height and width of each area, $h_{i}$,$1\leq{h_i}\leq{H}$ and $w_{i}$,$1\leq{w_i}\leq{W}$, as the features of the area. To combine these features, we use a multi-layer perceptron. To do so, we treat $h_{i}$ and $w_{i}$ as discrete values and project them onto a vector space using embedding (see Equation \ref{height_embed} and \ref{width_embed}).

\begin{equation}
\label{height_embed}
    e_i^{h}=\mathbbm{1}(h_i)E^h
\end{equation}

\begin{equation}
\label{width_embed}
    e_i^{w}=\mathbbm{1}(w_i)E^w
\end{equation}

where $\mathbbm{1}(h_i)$ and $\mathbbm{1}(w_i)$ are the one-hot encoding of $h_i$ and $w_i$, and $E^h\in{R^{H\times{S}}}$ and $E^w\in{R^{W\times{S}}}$ are the embedding matrices. $S$ is the depth of the embedding. We concatenate them to form the representation of the shape of an area.

\begin{equation}
\label{shape_embed}
    e_i=[e_i^{h}, e_i^{w}]
\end{equation}



We then combine them using a single-layer perceptron followed by a linear transformation (see Equation \ref{combine}).

\begin{equation}
\label{combine}
    k_i^r=\phi(\mu_{i}W_{\mu} + \sigma_{i}W_{\sigma} + e_{i}W_{e})W_{d}
\end{equation}

where $\phi$ is a nonlinear transformation such as ReLU, and $W_{\mu}\in{\mathbb{R}^{D\times{D}}}$, $W_{\sigma}\in{\mathbb{R}^{D\times{D}}}$, $W_{e}\in{\mathbb{R}^{2S\times{D}}}$ and $W_d\in{\mathbb{R}^{D\times{D}}}$. $W_{\mu}$, $W_{\sigma}$, $W_{e}$ and $W_d$ are trainable parameters.

\subsection{Fast Computation Using Summed Area Table}

If we naively compute $\mu_{i}$, $\sigma_{i}$ and $v_i^{r_i}$, the time complexity for computing attention will be $O(|M|A^2)$ where $|M|$ is the size of the memory that is $L$ for a 1-dimensional sequence or $L_vL_h$ for a 2-dimensional memory. $A$ is the maximum size of an area, which is $S$ in the one dimensional case and $WH$ in the 2-dimensional case. This is computationally expensive in comparison to the attention computed on the original memory, which is $O(|M|)$. To address the issue, we use summed area table, an optimization technique that has been used in computer vision for computing features on image areas \citep{Viola01rapidobject}. It allows constant time to calculate a summation-based feature in each rectangular area, which allows us to bring down the time complexity to $O(|M|A)$---We will report on the actual time cost in our experimental section.

Summed area table is based on an integral image \citep{Szeliski:2010:CVA:1941882}, $I$, which can be efficiently computed in a single pass of the memory. With the integral image, we can calculate the key and value of each area in constant time. 
We present the Pseudo code for performing Eq.~\ref{area_key_mean}, \ref{area_value} and \ref{area_var} as well as the shape size of each area in Algorithm~\ref{alg:main1} and ~\ref{alg:main2}. These Pseudo code are designed based on highly efficient Tensor operations~\footnote{See TensorFlow implementation of Area Attention as well as its integration with Transformer and LSTM in https://github.com/tensorflow/tensor2tensor.}.

\begin{algorithm}[t]
\caption{Compute the vector sum and the size of each area, for all the qualified rectangular areas on a given grid.}
 \label{alg:main1}
 \KwIn{A tensor $G$ in shape of $[H, W, D]$ that represents a grid with height $H$ and width $W$ where each item is a vector of depth $D$.}
 \KwOut{Sum of vectors of each area, $U$, and height and width of each area, $S_{h}$ and $S_{w}$.}

{\nonl{\bf{Hyperparameter}}: maximum allowed area width $W_{a}$ \\ \hskip 0.5cm and height $H_{a}$.}

{\bf Compute} horizontal integral image $I_{h}$ by cumulative sum \\ \hskip 0.5cm along horizontal direction over $G$\; 
{\bf Compute} integral image $I_{hv}$ by cumulative sum along \\ \hskip 0.5cm vertical directions over $I_{h}$\;
{\bf Compute} $I$ by padding all-zero vectors to the left and \\ \hskip 0.5cm top of $I_{hv}$\;
\For{$h=1,\cdots, H_{a}$}{
\For{$w=1,\cdots, W_{a}$}{
   $I_1 \leftarrow I[h+1:,w+1:,:]$ \;
   $I_2 \leftarrow I[:-h-1,:-w-1,:]$ \;
   $I_3 \leftarrow I[h+1:,:-w-1,:]$ \;
   $I_4 \leftarrow I[:-h-1,w+1:,:]$ \;
   $\bar{U}=I_1+I_2-I_3-I_4$ \;
   $\bar{S_{h}} \leftarrow [h]^{(H-h)\times{(W-w)}}$; Fill tensor with value $h$ for the height of each area\;
    $\bar{S_{w}} \leftarrow [w]^{(H-h)\times{(W-w)}}$; Fill tensor with value $w$ for the width of each area\;
   $S_{h}\leftarrow [S_{h}\ \ \bar{S_{h}}]$, reshape $\bar{S_{h}}$ to $[-1, 1]$ and concatenate on the first dimension\;
   $S_{w}\leftarrow [S_{w}\ \ \bar{S_{w}}]$, reshape $\bar{S_{w}}$ to $[-1, 1]$ and concatenate on the first dimension\;
   $U\leftarrow [U\ \ \bar{U}]$, reshape $\bar{U}$ to $[-1, D]$ and concatenate on the first dimension\;
 }
 }
 \KwRet $U$, $S_{h}$ and $S_{w}$.
\end{algorithm}

\begin{algorithm}[t]
\caption{Compute the vector mean, standard deviation, and sum as well as the size of each area, for all the qualified rectangular areas on a grid.}
 \label{alg:main2}
 \KwIn{A tensor $G$ in shape of $[H, W, D]$ that represents a grid with height $H$ and width $W$ where each item is a vector of depth $D$.}
 \KwOut{Vector mean $\mu$, standard deviation $\sigma$ and sum $U$ as well as height $S_{h}$ and width $S_{w}$ of each area.}
 Acquire $U$, $S_{h}$ and $S_{w}$ using Algorithm \ref{alg:main1} with input $G$\;
 Acquire $U^{'}$ using Algorithm \ref{alg:main1} with input $G \odot G$ where $\odot$ is for element-wise multiplication\;
 $\mu \leftarrow U \oslash {S}$ where $\oslash$ is for element-wise division\;
 $\mu^{'} \leftarrow U^{'} \oslash {S}$ \;
 $\sigma \leftarrow \sqrt{\mu^{'}- \mu \odot \mu}$ \;
 \KwRet $\mu$, $\sigma$, $U$, as well as $S_{h}$ and $S_{w}$.
\end{algorithm}

\section{Experiments}
We experimented with area attention on two important tasks: neural machine translation (including both token and character-level translation) and image captioning, where attention mechanisms have been a common component in model architectures for these tasks. The architectures we investigate involves several popular encoder and decoder choices, such as LSTM \citep{Hochreiter:1997:LSM:1246443.1246450} and Transformer \citep{DBLP:journals/corr/VaswaniSPUJGKP17}. The attention mechansims in these tasks include both self attention and encoder-decoder attention. Note that area attention does not change the size of queries and only expands the size of keys and values. The basic form of area attention is completely parameter free. As a result, all the models in Table 1-5 using (Eq.\ref{area_key_mean} \& \ref{area_value}) have the same number of parameters as the corresponding baseline models, which allows a fair comparison. 

\subsection{Token-Level Neural Machine Translation}
Transformer has recently \citep{DBLP:journals/corr/VaswaniSPUJGKP17} established the state of art performance on WMT 2014 English-to-German and English-to-French tasks, while LSTM with encoder-decoder attention has been a popular choice for neural machine translation tasks. We use the same dataset as the one used in \citep{DBLP:journals/corr/VaswaniSPUJGKP17} in which the WMT 2014 English-German (EN-DE) dataset contains about 4.5 million English-German sentence pairs, and the English-French (EN-FR) dataset has about 36 million English-French sentence pairs \citep{DBLP:journals/corr/WuSCLNMKCGMKSJL16}. A token is either a byte pair \citep{DBLP:journals/corr/BritzGLL17} or a word piece \citep{DBLP:journals/corr/WuSCLNMKCGMKSJL16} as in the original Transformer experiments. We performed three runs for each configuration and report the average of these runs. * stands for statistical significance ($p<0.05$) for comparison with regular attention and ** for statistical significance when comparing with all the other model conditions.

\subsubsection{Transformer Token-Level MT Experiments} 

Transformer heavily uses attentional mechanisms, including both self-attention in the encoder and the decoder, and attention from the decoder to the encoder. We vary the configuration of Transformer to investigate how area attention impacts the model. In particular, we investigated the following variations of Transformer: \textit{Tiny} (\#hidden layers=2, hidden size=128, filter size=512, \#attention heads=4), \textit{Small} (\#hidden layers=2, hidden size=256, filter size=1024, \#attention heads=4), \textit{Base} (\#hidden layers=6, hidden size=512, filter size=2048, \#attention heads=8) and \textit{Big} (\#hidden layers=6, hidden size=1024, filter size=4096 for EN-DE and 8192 for EN-FR, \#attention heads=16).

During training, sentence pairs were batched together based on their approximate sequence lengths. All the model variations except \textit{Big} uses a training batch contained a set of sentence pairs that amount to approximately 32,000 source and target tokens and were trained on one machine with 8 NVIDIA P100 GPUs for a total of 250,000 steps. Given the batch size, each training step for the Transformer Base model, on 8 NVIDIA P100 GPUs, took 0.4 seconds for Regular Attention, 0.5 seconds for the basic form of Area Attention (Eq.\ref{area_key_mean} and Eq.\ref{area_value}), 0.8 seconds for Area Attention using multiple features (Eq.\ref{combine} and Eq.\ref{area_value}).

For \textit{Big}, due to the memory constraint, we had to use a smaller batch size that amounts to roughly 16,000 source and target tokens and trained the model for 600,000 steps. Each training step took 0.5 seconds for Regular Attention, 0.6 seconds for the basic form of Area Attention (Eq.\ref{area_key_mean} and \ref{area_value}), 1.0 seconds for Area Attention using multiple features (Eq.\ref{combine} and \ref{area_value}). Similar to previous work, we used the Adam optimizer with a varying learning rate over the course of training---see \citep{DBLP:journals/corr/VaswaniSPUJGKP17} for details. 


\bgroup
\def\arraystretch{1.2}
\begin{table*}[h!]
\caption{The BLEU scores on token-level translation tasks for the variations of the Transformer-based architecture.}
\label{transformer-token-table}
\centering
\begin{tabular}{lllllllllll}
\toprule
\multicolumn{1}{c}{\multirow{2}{*}{Model}} & \multicolumn{2}{c}{Regular Attention} & & \multicolumn{3}{c}{Area Attention (Eq.\ref{area_key_mean} and \ref{area_value})} & & \multicolumn{3}{c}{Area Attention (Eq.\ref{combine} and \ref{area_value})} \\
\cline{2-3}\cline{5-7}\cline{9-11}
\multicolumn{1}{c}{} & EN-DE & EN-FR & & & EN-DE & EN-FR & & & EN-DE & EN-FR\\
\midrule
Tiny & 18.58 & 27.03 & & & $19.13^{*}$ & $27.4^{*}$ & & & $\textbf{19.27}^{*}$ & $\textbf{27.91}^{**}$ \\
Small & 22.55 & 31.93 & & & 22.84 & $32.31^{*}$ & & & $\textbf{23.2}^{**}$ & $\textbf{32.93}^{**}$ \\
Base & 28.16 & 38.97 & & & 28.47 & $\textbf{39.27}^*$ & & & $\textbf{28.52}^{*}$ & 39.19 \\
Big & 29.26 & 41.0 & & & 29.49 & 41.18 & & & $\textbf{29.77}^{*}$ & $\textbf{41.46}^{*}$ \\
\bottomrule
\end{tabular}
\end{table*}
\egroup


We applied area attention to each of the Transformer variation to both encoder and decoder self-attention, and the encoder-decoder attention in the first two layers. We found area attention consistently improved Transformer on all the model variations (see Table \ref{transformer-token-table}), even with the basic form of area attention where no additional parameters are used (Eq.\ref{area_key_mean} and Eq.\ref{area_value}). For Transformer Base, area attention achieved BLEU scores (EN-DE: 28.52 and EN-FR: 39.27) that surpassed the previous results for both EN-DE and EN-FR.

For EN-FR, the performance of Transformer Big with regular attention---a baseline---does not match what was reported in the Transformer paper~\citep{DBLP:journals/corr/VaswaniSPUJGKP17}, largely due to a different batch size and the different number of training steps used, although area attention still outperformed the baseline consistently. On the other hand, area attention with Transformer Big achieved BLEU \textit{29.77} on EN-DE that improved upon the state-of-art result of 28.4 reported in \citep{DBLP:journals/corr/VaswaniSPUJGKP17} with a significant margin.

\bgroup
\def\arraystretch{1.2}
\begin{table*}[h!]
\caption{The BLEU scores on token-level translation tasks for the LSTM-based architecture with varying model capacities.}
\label{lstm-token-table}
\centering
\begin{tabular}{llllllllllll}
\toprule
 \multicolumn{1}{c}{\multirow{2}{*}{\#Cells}} &
 \multicolumn{1}{c}{\multirow{2}{*}{\#Heads}} &
 \multicolumn{2}{c}{Regular Attention} & & \multicolumn{3}{c}{Area Attention (Eq.\ref{area_key_mean},\ref{area_value})} & & \multicolumn{3}{c}{Area Attention (Eq.\ref{combine},\ref{area_value})}\\
\cline{3-4}\cline{6-8}\cline{10-12}
\multicolumn{1}{c}{} & & EN-DE & EN-FR & & & EN-DE & EN-FR & & & EN-DE & EN-FR \\
\midrule
256 & 1 & 16.58 & 22.77 & & & $19.26^{*}$ & $29.35^{*}$ & & &
$\textbf{19.46}^{*}$ & $\textbf{29.79}^{**}$ \\
256 & 4 & 16.73 & 28.1 & & & $20.25^{*}$ & $\textbf{30.49}^{*}$ & & & $\textbf{20.74}^{**}$ & $30.2^{*}$ \\
512 & 1 & 18.65 & 30.32 & & & $\textbf{21.82}^{*}$ & $\textbf{32.80}^{*}$ & & & $21.80^{*}$ & $32.73^{*}$ \\
512 & 4 & 19.16 & 30.55 & & & $23.09^{*}$ & $33.75^{*}$ & & & $\textbf{23.41}^{*}$ & $\textbf{34.09}^{**}$ \\
1024 & 1 & 19.4 & 31.99 & & & $\textbf{23.69}^{*}$ & $34.65^{*}$ & & & $23.48^{*}$ & $\textbf{34.76}^{*}$ \\
1024 & 4 & 20.21 & 32.21 & & & $24.55^{*}$ & $35.95^{*}$ & & & $\textbf{24.85}^{**}$ & $\textbf{35.97}^{*}$ \\
\bottomrule
\end{tabular}
\end{table*}
\egroup

\subsubsection{LSTM Token-Level MT Experiments} 
We used a 2-layer LSTM for both encoder and decoder. The encoder-decoder attention is based on multiplicative attention where the alignment of a query and a memory key is computed as their dot product \citep{DBLP:journals/corr/LuongPM15}. We vary the size of LSTM and the number of attention heads to investigate how area attention can improve LSTM with varying capacity on translation tasks. The purpose is to observe the impact of area attention on each LSTM configuration, rather than for a comparison with Transformer. 

Because LSTM requires sequential computation along a sequence, it trains rather slow compared to Transformer. To improve GPU utilization we increased data parallelism by using a much larger batch size than training Transformer. We trained each LSTM model on one machine with 8 NVIDIA P100. For a model has 256 or 512 LSTM cells, we trained it for 50,000 steps using a batch size that amounts to approximately 164,000 source and target tokens. When the number of cells is 1024, we had to use a smaller batch size with roughly 131,000 tokens, due to the memory constraint, and trained the model for 625,000 steps. 

In these experiments, we used areas of maximum size 2 and attention was computed from the output of the decoder's top layer to that of the encoder. Similar to what we observed with Transformer, area attention consistently improves LSTM architectures in all cases (see Table~\ref{lstm-token-table}). 

\subsection{Character-Level Neural Machine Translation}

Compared to token-level translation, character-level translation requires the model to address significantly longer sequences, which are a more difficult task and often less studied. We speculate that the ability to combine adjacent characters, as enabled by area attention, is likely useful to improve a regular attentional mechanisms. Likewise, we experimented with the same set of Transformer and LSTM-based architectures for this task (see the appendix for experimental details). 

Transformer has not been used for character-level translation tasks. We found area attention consistently improved Transformer across all the model configurations. The best result we found in the literature is $BLEU=23.75$ in \citep{Bytenet2017} and next $BLEU=22.62$ reported by \citep{DBLP:journals/corr/WuSCLNMKCGMKSJL16}. We achieved $BLEU=26.57$ for the English-to-German character-level translation task and $BLEU=35.16$ on the English-to-French character-level translation task. Note that these accuracy gains are based on the basic form of area attention (see Eq.\ref{area_key_mean} and Eq.\ref{area_value}), which does not add any trainable parameters to the model. 

Similarly, we tested LSTM architectures on the character-level translation tasks. We found area attention outperformed the baselines in all the conditions (see Table \ref{lstm-char-table}).

\bgroup
\def\arraystretch{1.2}
\begin{table}[h!]
\caption{The BLEU scores on character-level translation tasks for the Transformer-based architecture with varying model capacities.\label{tbl_app}}
\label{transformer-table}
\centering
\begin{tabular}{lllllll}
\toprule
\multicolumn{1}{c}{\multirow{2}{*}{Model}} & \multicolumn{2}{c}{Regular} & & \multicolumn{3}{c}{Area (Eq.\ref{area_key_mean}, \ref{area_value})} \\
\cline{2-3}\cline{5-7}
\multicolumn{1}{c}{} & EN-DE & EN-FR & & & EN-DE & EN-FR \\
\midrule
Tiny & 6.9 & 9.58 & & & \textbf{8.2} & $\textbf{11.69}^{*}$ \\
Small & 12.16 & 18.43 & & & $\textbf{13.46}^{*}$ & $\textbf{21.05}^{*}$ \\
Base & 24.88 & 33.26 & & & \textbf{24.96} & \textbf{33.68} \\
Big & 25.96 & 33.82 & & & \textbf{26.57} & \textbf{35.16} \\
\bottomrule
\end{tabular}
\end{table}
\egroup





\bgroup
\def\arraystretch{1.2}
\begin{table}[h!]
\caption{The BLEU scores on character-level translation tasks for the LSTM-based architecture with varying model capacities.}
\label{lstm-char-table}
\centering
\begin{tabular}{lllllll}
\toprule
 \multicolumn{1}{c}{\multirow{2}{*}{Cell,Head}} &
 \multicolumn{2}{c}{Regular} & & \multicolumn{3}{c}{Area (Eq.\ref{area_key_mean}, \ref{area_value})} \\
\cline{2-3}\cline{5-7}
\multicolumn{1}{c}{} & EN-DE & EN-FR & & & EN-DE & EN-FR \\
\midrule
256, 1 & 9.9 & 16.82 & & & \textbf{11.23} & \textbf{17.27} \\
256, 4 & 11.99 & 17.36 & & & \textbf{12.51} & $\textbf{19.85}^{*}$ \\
512, 1 & 16.37 & 23.7 & & & \textbf{16.86} & \textbf{24.13} \\
512, 4 & 16.43 & 24.71 & & & \textbf{17.61} & \textbf{25.36} \\
1024, 1 & 20.26 & 27.83 & & & $\textbf{21.05}^{*}$ & \textbf{29.48} \\
1024, 4 & 21.28 & 28.53 & & & \textbf{21.71} & $\textbf{30.68}^{*}$ \\
\bottomrule
\end{tabular}
\end{table}
\egroup

\bgroup
\def\arraystretch{1.2}
\begin{table*}[t]
\caption{Test accuracy of image captioning models on COCO40 (in-domain) and Flickr 1K (out-of-domain) tasks.\label{caption}}
\label{transformer-table2}
\centering
\begin{tabular}{lllllll}
\toprule
\multicolumn{1}{c}{\multirow{2}{*}{Model}} & \multicolumn{2}{c}{COCO40} & & \multicolumn{3}{c}{Flickr 1K} \\
\cline{2-3}\cline{5-7}
\multicolumn{1}{c}{} & CIDEr & ROUGE-L & & & CIDEr & ROUGE-L \\
\midrule
Benchmark \citep{DBLP:conf/acl/SoricutDSG18} & 1.032 & 0.700 & & & 0.359 & 0.416 \\
Benchmark Replicate & 1.034 & 0.701 & & & 0.355 & 0.409 \\
$2\times{2}$ Eq.\ref{area_key_mean} \& \ref{area_value} & \textbf{1.060} & 0.704 & & & 0.364 & \textbf{0.420} \\
$3\times{3}$ Eq.\ref{area_key_mean} \& \ref{area_value} & \textbf{1.060} & 0.706 & & & \textbf{0.377} & 0.419 \\
$3\times{3}$ Eq.\ref{combine} \& \ref{area_value} & 1.045 & \textbf{0.707} & & & 0.372 & \textbf{0.420} \\
\bottomrule
\end{tabular}
\end{table*}
\egroup

\subsection{Image Captioning}

Image captioning is the task to generate natural language description of an image that reflects the visual content of an image. This task has been addressed previously using a deep architecture that features an image encoder and a language decoder \citep{DBLP:journals/corr/XuBKCCSZB15,DBLP:conf/acl/SoricutDSG18}. The image encoder typically employs a convolutional net such as ResNet \citep{DBLP:journals/corr/HeZRS15} to embed the images and then uses a recurrent net such as LSTM or Transformer \citep{DBLP:conf/acl/SoricutDSG18} to encode the image based on these embeddings. For the decoder, either LSTM \citep{DBLP:journals/corr/XuBKCCSZB15} or Transformer \citep{DBLP:conf/acl/SoricutDSG18} has been used for generating natural language descriptions. In many of these designs, attention mechanisms have been an important component that allows the decoder to selectively focus on a specific part of the image at each step of decoding, which often leads to better captioning quality.

In this experiment, we follow a champion condition in the experimental setup of \citep{DBLP:conf/acl/SoricutDSG18} that achieved state-of-the-art results as our benchmark model. It uses a pre-trained Inception-ResNet to generate $8\times{8}$ image embeddings, a 6-layer Transformer for image encoding and a 6-layer Transformer for decoding. The benchmark model has a hidden size of 512 and uses 8-head regular attention. To investigate how area attention improves the captioning accuracy, particularly regarding self-attention and encoder-decoder attention computed off the image, we add area attention with different maximum area sizes to the first 2 layers of the image encoder self-attention and encoder-decoder (caption-to-image) attention (see Table \ref{caption}), which both resemble a 2-dimensional area attention case. $2\times{2}$ stands for the maximum area size 2 by 2 and $3\times{3}$ for 3 by 3. For the $2\times{2}$ case, an area can be 1 by 1, 2 by 1, 1 by 2, and 2 by 2 as illustrated in Figure \ref{area-2d}. $3\times{3}$ allows more area shapes.

Similar to \citep{DBLP:conf/acl/SoricutDSG18}, we trained each model based on the training \& development sets provided by the COCO dataset \citep{DBLP:journals/corr/LinMBHPRDZ14}, which as 82K images for training and 40K for validation. Each of these images have at least 5 groudtruth captions. The training was conducted on a distributed learning infrastructure \citep{Dean:2012:LSD:2999134.2999271} with 10 GPU cores where updates are applied asynchronously across multiple replicas. We then tested each model on the COCO40 \citep{DBLP:journals/corr/LinMBHPRDZ14} and the Flickr 1K \citep{DBLP:journals/corr/Young14} test sets. Flickr 1K is out-of-domain for the trained model. For each experiment, we report CIDEr \citep{DBLP:journals/corr/VedantamZP14a} and ROUGE-L \citep{Lin:2004:OME:1220355.1220427} metrics. For both metrics, higher number means better captioning accuracy---the closer distances between the predicted and the groundtruth captions. Similar to the previous work \citep{DBLP:conf/acl/SoricutDSG18}, we report the numerical values returned by the COCO online
evaluation server\footnote{http://mscoco.org/dataset/\#captions-eval} for the COCO C40 test set. Previous work \citep{DBLP:conf/acl/SoricutDSG18} has revealed that human evaluation would give a more complete examination about the model accuracy, which we leave out in this work as our focus is on area attention as a general mechanism. 

We found models with area attention outperformed the benchmark on both CIDEr and ROUGE-L metrics with a large margin (see Table \ref{caption}). The models with $2\times{2}$ Eq.\ref{area_key_mean} and $3\times{3}$ Eq.\ref{area_key_mean} do not use any additional parameters beyond the benchmark model. $3\times{3}$ achieved the best results overall. $3\times{3}$ Eq. \ref{combine} adds a small fraction of the number of parameters to the benchmark model and did not seem to improve on the parameter-free version of area attention, although it still outperformed the benchmark.

\section{Discussions}

In this paper, we focus on mean (Equation \ref{area_key_mean}) and sum pooling (Equation \ref{area_value}) as a way to compute keys and values for each area. As we can see from the experimental results, these simple parameter-free area representations can bring accuracy gain to a range of tasks. As mentioned earlier, it is possible to use other methods such as max pooling for this purpose. We experimented with max pooling on the Transformer model for both machine translation and image captioning tasks. For token-level translation, max pooling with Transformer Base achieved BLEU 28.48 for EN-DE and 39.21 for EN-FR. For character-level translation, max pooling with Transformer Base achieved 24.92 and 33.84, respectively. Similarly, for image captioning tasks, max pooling with a 2x2 maximum area size achieved CIDEr 1.055 and ROUGE-L 0.706 for COCO40 official tests, and CIDEr 0.365 and ROUGE-L 0.416 for Flickr 1K tests. While max pooling seems to offer comparable results on some of the tasks, the exact solution cannot be efficiently computed. Without using summed area table, iterating over each area is significantly slower. Alternatively, we can compute and pool over all the areas in parallel, But it requires significantly more memory that limits the use of a large batch size or the handling of a long sequence such as character-level translation. It is possible to calculate approximate max pooling based on summed area table \citep{approximate_max}, although it can incur numerical problems such as underflow. To handle very long sequences, we can apply area attention to a neighborhood of a sequence given a query instead of the entire sequence. These ideas deserve further investigation.

We found the benefit of area attention is more pronounced when a model is relatively small (see Table 1-4). It also appears that the improvement for LSTM-based architectures is quite substantial particularly on token-level translation tasks, e.g., Table 2. For character-level translation tasks, although the improvement from area attention is quite consistent across model conditions, there is not as much statistical significance as we acquired for token-level translation. One reason is that there is a larger variance in the results from which we speculate that more training iterations are needed for better convergence.

\begin{figure}[h!]
    \centering
    \label{att_viz}
    \begin{subfigure}{0.5\textwidth}
        
        \centering
        \includegraphics[width=0.475\linewidth]{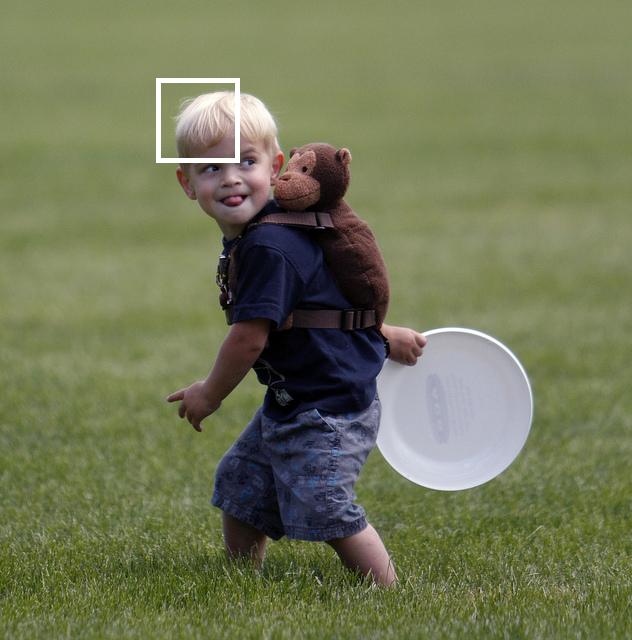}%
        \hfill
        \includegraphics[width=0.475\linewidth]{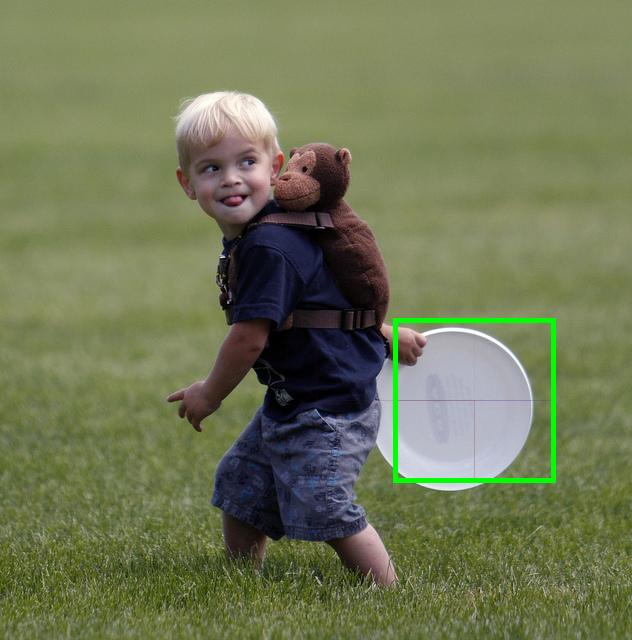}
        
        \caption{a young boy holding a frisbee in his hand }
    \end{subfigure}
     \vskip\baselineskip 
    \begin{subfigure}{0.5\textwidth}
    \label{att_viz2}
        \centering
        \includegraphics[width=0.475\linewidth]{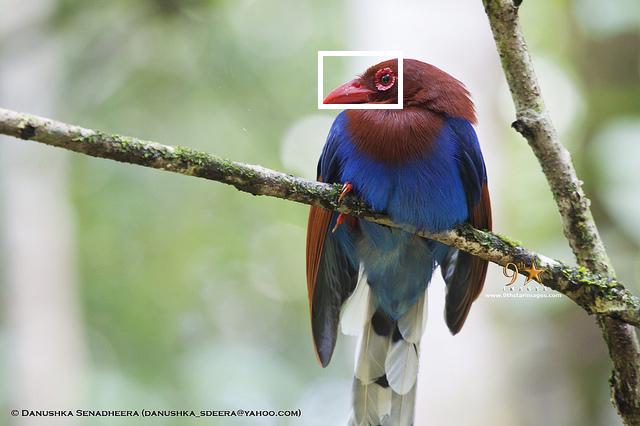}%
        \hfill
        \includegraphics[width=0.475\linewidth]{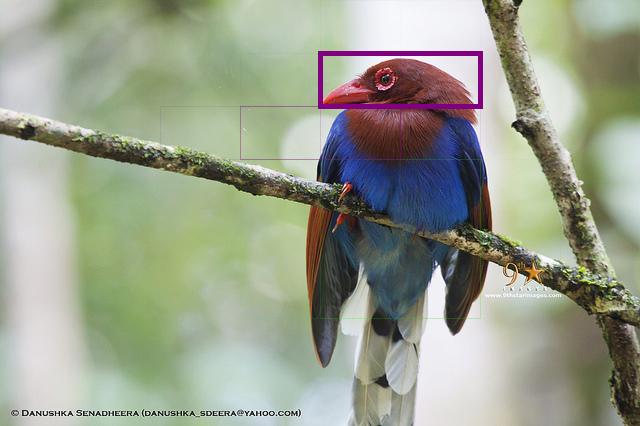}
        
        \caption{a colorful bird perched on a branch of a tree}
    \end{subfigure}
     \vskip\baselineskip
    \begin{subfigure}{0.5\textwidth}
    \label{att_viz3}
        \centering
        \includegraphics[width=0.475\linewidth]{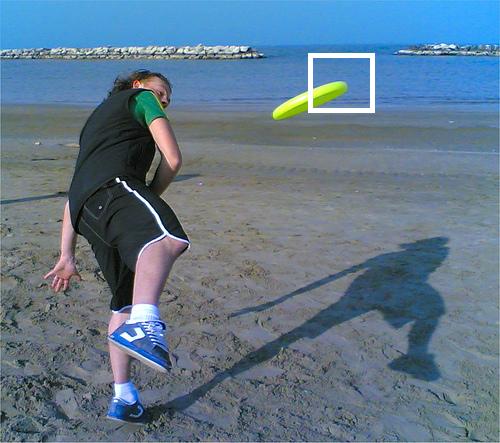}%
        \hfill
        \includegraphics[width=0.475\linewidth]{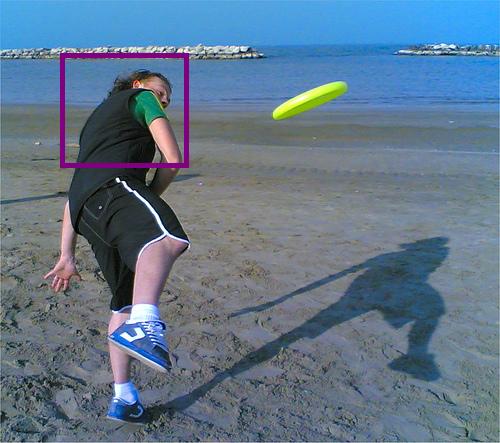}
        \caption{a man on a beach throwing a frisbee}
    \end{subfigure}
     \vskip\baselineskip 
    \caption{The images on the left show the query grid and the images on the right are attended areas. The top attended area is highlighted with the color of the corresponding attention head.}
    \label{fig:imageca}
\end{figure}

In addition to offering better accuracy, we want to better understand how area attention works, particularly regarding if area attention is able to capture structural or semantic coherence in the data. To do so, we analyze the learned multi-head area self-attention in Transformer encoder for the image captioning task (see examples in Figure \ref{fig:imageca} and additional examples in the appendix). From these examples, we can see area attention often appropriately captures the image areas that are relevant to the query grid. In particular, many of the top-attended areas (shown in bold) include more than one grid with a variety of shapes, depending on the scene. 

Similarly, we have analyzed the self-attention for the character-level machine translation tasks (see examples in the appendix). The analysis reveals that area attention enables multi-head attention in Transformer to attend to the whole word that the query character belongs to as well as other relevant words in the sentences. This shows that area attention allows the model to attend to appropriate granularity of information that is more consistent with the structural and semantics coherence in the data.



\section{Conclusions}
In this paper, we present a novel attentional mechanism by allowing the model to attend to areas as a whole. An area contains one or a group of items in the memory to be attended. The items in the area are either spatially adjacent when the memory has 2-dimensional structure, such as images, or temporally adjacent for 1-dimensional memory, such as natural language sentences. Importantly, the size of an area, i.e., the number of items in an area or the level of aggregation, can vary depending on the learned coherence of the adjacent items, which gives the model the ability to attend to information at varying granularity. Area attention contrasts with the existing attentional mechanisms that are item-based. We evaluated area attention on two tasks: neural machine translation and image captioning, based on model architectures such as Transformer and LSTM. Area attention is able to offer further improvement on accuracy consistently across a variety of tasks over these strong baselines.

\section*{Acknowledgements}
We would like to thank the anonymous reviewers for their insightful feedback that substantially improved the paper. We also want to thank the readers of the early versions of the paper for their constructive comments.

\bibliography{area_attention}
\bibliographystyle{icml2019}

\newpage

\end{document}


\maketitle

\section{Experimental Details}
For both token and character-level translation tasks, we used a maximum area size of 5 for the two layers for Transformer Tiny and Small, and a maximum area size of 4 for the first layer of Transformer Base with Eq.3, and for the first two layers of Transformer Base with Eq.9. For Transformer Big EN-DE, we used a maximum area size of 4 for the first layer with Eq.3, and a maximum area size of 3 for Eq.9. For Transformer Big EN-FR, we used a maximum area size of 3 for the first two layers with Eq.3, and a maximum area size of 4 for the first layer with Eq.9.

For character-level translation tasks, we here used the same dataset, and the experimental strategies as the token-level experiments (see Section 4.1.1) with a few differences to reduce the experiment time because it is significantly slower to train than token-level translation. In particular, we trained all the Transformer Big for 300,000 steps. For Transformer Big EN-DE, we could use a larger batch size that amounts to approximately 32,000 characters. All the LSTM models use the same batch size that amounts to 164,000 characters for 50,000 steps.

\section{Additional Experimental Results}

We evaluated the approach of area feature combination (Eq.9) on character-level translation tasks as well. It performed on par with the basic form of area attention (Eq.3). In particular, it outperformed the basic form (see details in Table 3), with statistical signficance, on Transformer Tiny EN-FR ($BLEU=12.91$) and Transformer Small EN-FR ($BLEU=21.93$) and EN-DE ($BLEU=14.5$), which seem to imply that when the basic area attention helps, the method of area feature combination could bring further improvements.

We also explored the approach of using normalized sigmoid (Shen & Lee, 2016; Rei & Søgaard, 2018) as the activation function of multi-head attention in Transformer. The quick experiments with Transformer Tiny and Small by replacing softmax with normalized sigmoid led to poor results, which deserves further investigation.

\section{Related Algorithmic Details for Integral Images}

Summed area table is based on an integral image, $I$, which can be efficiently computed in a single pass of the memory (see Equation \ref{eq:integral}). Here let us focus on the area value calculation for a 2-dimensional memory because a 1-dimensional memory is just a special case with the height of the memory grid as 1.

\begin{equation}
\label{eq:integral}
    I_{x,y} =v_{x,y} + I_{x,y-1} + I_{x-1,y} - I_{x-1, y-1}
\end{equation}

where $x$ and $y$ are the coordinates of the item in the memory. With the integral image, we can calculate the key and value of each area in constant time. The sum of all the vectors in a rectangular area can be easily computed as the following (Equation \ref{eq:sum}).

\begin{equation}
    v_{x1,y1,x2,y2} = I_{x2,y2} + I_{x1,y1} - I_{x2,y1} - I_{x1,y2}
    \label{eq:sum}
\end{equation}

where $v_{x1,y1,x2,y2}$ is the value for the area located with the top-left corner at $(x_1,y_1)$ and the bottom-right corner at $(x_2,y_2)$. By dividing $v_{x1,y1,x2,y2}$ with the size of the area, we can easily compute $\mu_{x1,y1,x2,y2}$. Based on the summed area table, $\sigma_{x1,y1,x2,y2}^{2}$ (thus $\sigma_{x1,y1,x2,y2}$) can also be computed at constant time for each area (see Equation \ref{quick_sigma}), where $I_{x,y}^2 =v_{x,y}^2 + I_{x,y-1}^2 + I_{x-1,y}^2-I_{x-1,y-1}^2$, which is the integral image of the element-wise squared memory.

\begin{equation}
\label{quick_sigma}
    \sigma_{x1,y1,x2,y2}^{2} = \dfrac{I_{x2,y2}^2 + I_{x1,y1}^2 - I_{x2,y1}^2 - I_{x1,y2}^2}{(x2-x1)\times{(y2-y1)}}-\mu_{x1,y1,x2,y2}^2
\end{equation}

The core component for computing these quantities is to be able to quickly compute the sum of vectors in each area after we obtain the integral image table $I$ for each coordinate $[x,y]$, as shown in Equation \ref{eq:integral} and \ref{eq:sum}. The Pseudo code for performing these are presented in the paper, which is based on efficient Tensor operations (see Algorithm 1 and 2).

\section{Attention Visualization}
To understand the behavior of area attention, we analyzed the attention distribution of the trained Transformer models for both image captioning (see Figure 1) and character-level machine translation tasks (see Figure 2). For both visualizations, we analyzed the encoder self attention. For image captioning, it is the relationship between a fixed-sized query cell from a 8x8 grid to a varying-sized area that can involve multiple adjacent grids on an image. For character-level machine translation, it shows how a query character attends to a group of adjacent characters in the same sentence. 

\begin{figure}[h]
    \centering
   \begin{adjustbox}{minipage=\linewidth,scale=0.70}
    \begin{subfigure}{0.5\textwidth}
        \centering
        \includegraphics[width=0.475\linewidth]{query_COCO_val2014_000000100489.jpg}%
        \hfill
        \includegraphics[width=0.475\linewidth]{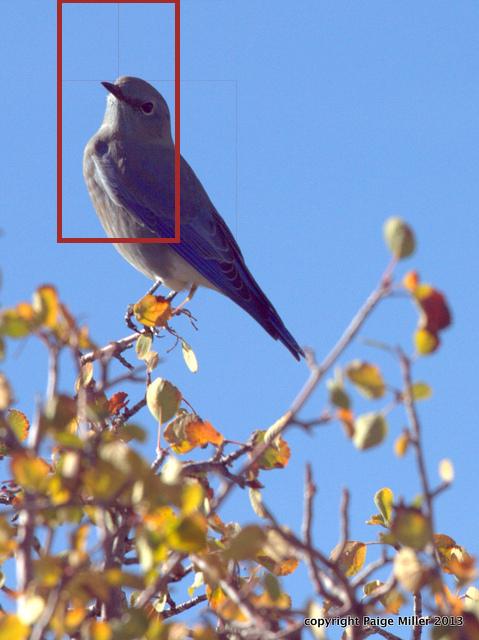}
        \caption{a bird sitting on top of a tree branch}
      \end{subfigure}
    \vskip\baselineskip \vspace{-10pt}
    \begin{subfigure}{\textwidth}
        \centering
        \includegraphics[width=0.475\linewidth]{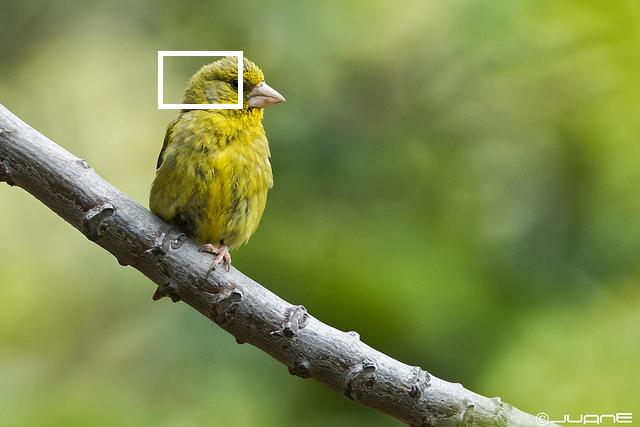}%
        \hfill
        \includegraphics[width=0.475\linewidth]{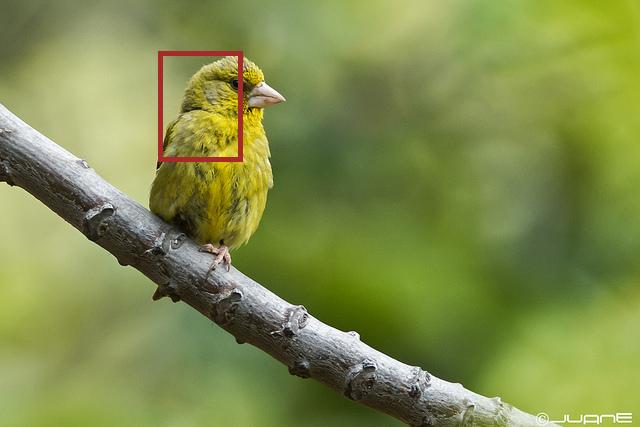}
        \caption{a small yellow bird perched on a branch}
    \end{subfigure}
    \vskip\baselineskip \vspace{-10pt}
    \begin{subfigure}{\textwidth}
    \label{att_viz4}
        \centering
        \includegraphics[width=0.475\linewidth]{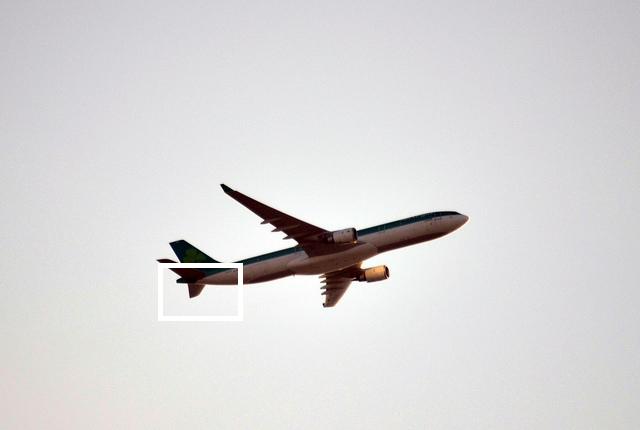}%
        \hfill
        \includegraphics[width=0.475\linewidth]{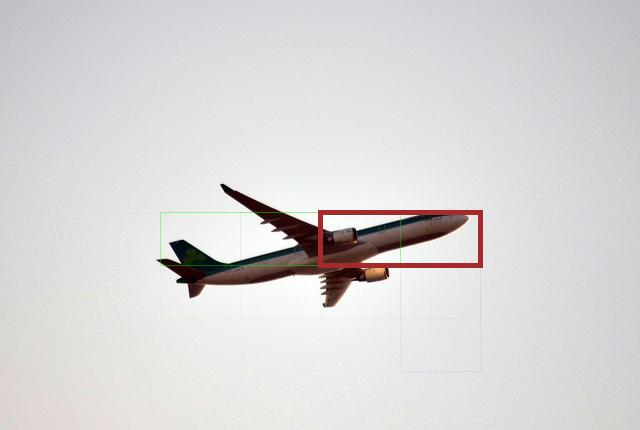}
        \caption{an airplane is flying in the sky}
    \end{subfigure}
    \vskip\baselineskip \vspace{-10pt}
    \begin{subfigure}{\textwidth}
        \centering
        \includegraphics[width=0.475\linewidth]{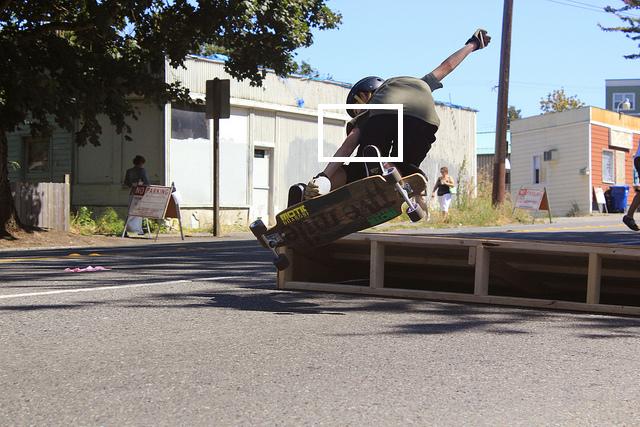}%
        \hfill
        \includegraphics[width=0.475\linewidth]{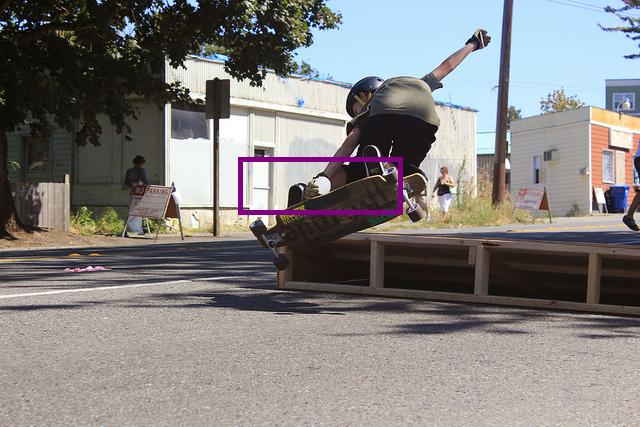}
        \caption{a man flying through the air while riding a skateboard}
    \end{subfigure}
    \vskip\baselineskip \vspace{-10pt}
    \caption{The images on the left show the query grid and the images on the right are attended areas. The top attended area is highlighted with the color of the corresponding attention head.}
    \end{adjustbox}
\end{figure}

\vspace{30pt}
\begin{figure}[h]
    \centering
    \begin{adjustbox}{minipage=\linewidth,scale=0.75}
    \begin{subfigure}{\textwidth}
        \centering
        \includegraphics[width=1.1\linewidth]{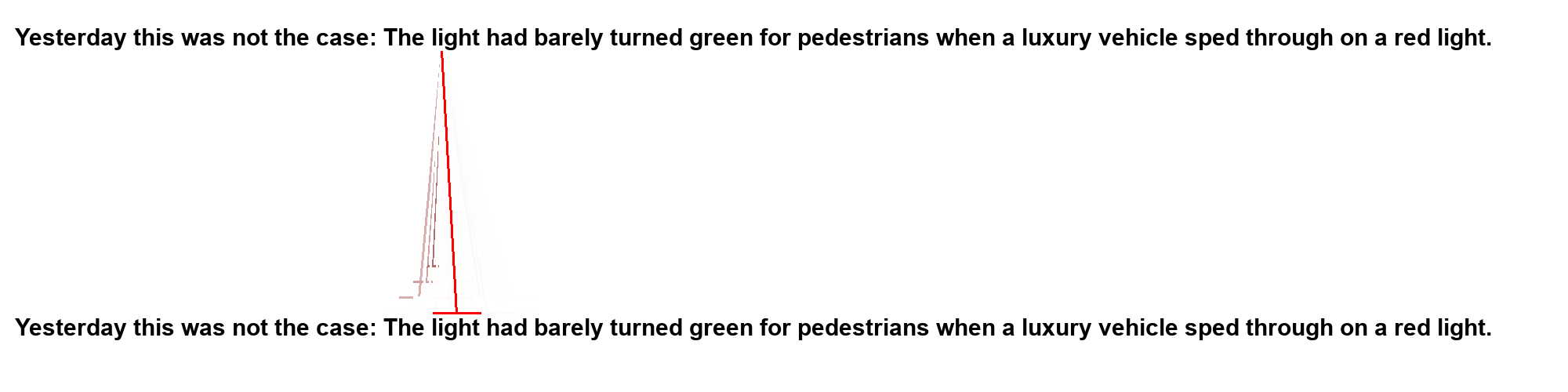}%
        \caption{}
      \end{subfigure}
      \begin{subfigure}{\textwidth}
       \centering
      \includegraphics[width=1.1\linewidth]{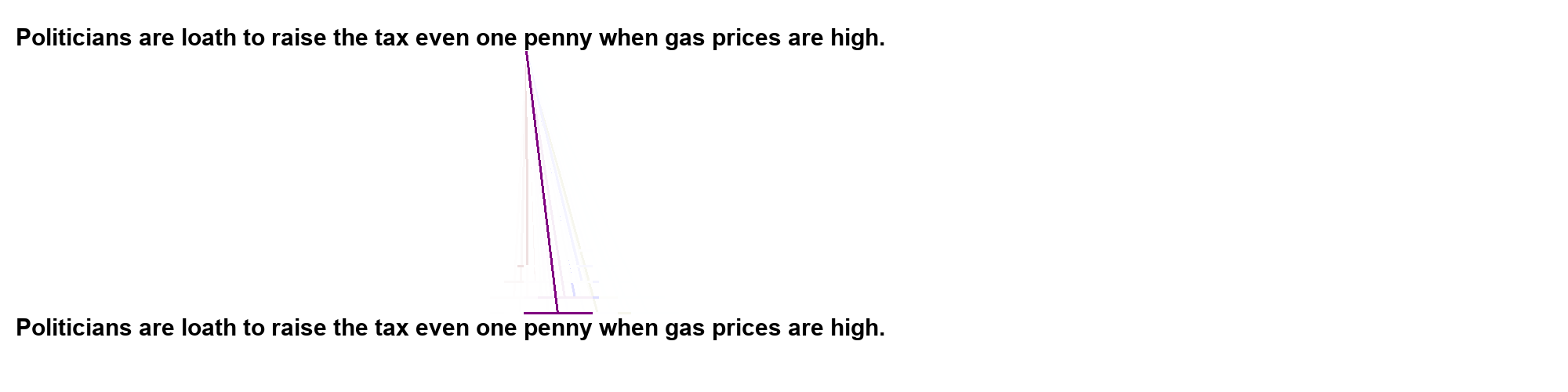}
      \caption{}
      \end{subfigure}
            \begin{subfigure}{\textwidth}
       \centering
      \includegraphics[width=1.1\linewidth]{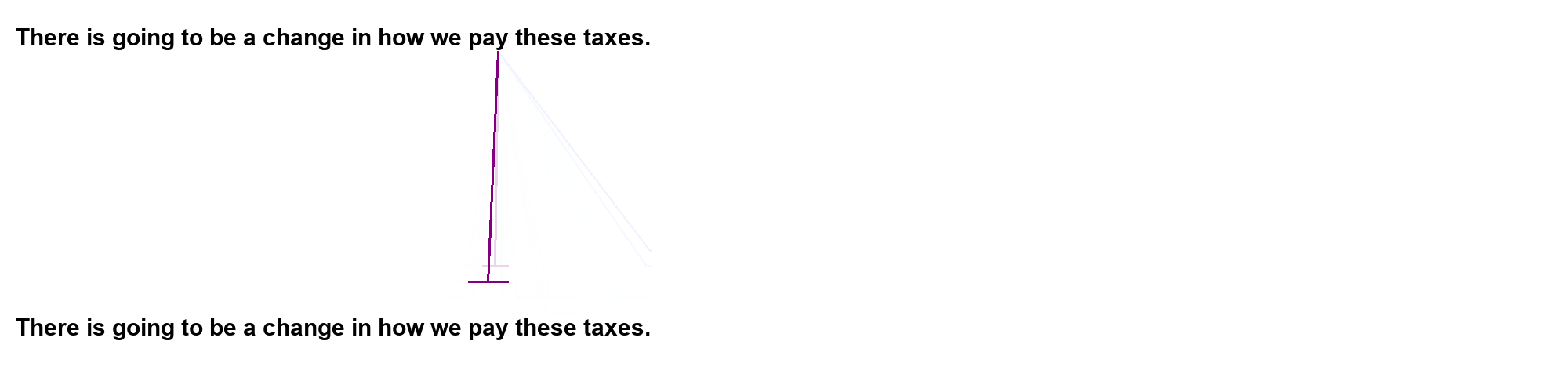}
      \caption{}
      \end{subfigure}
            \begin{subfigure}{\textwidth}
      \centering
      \includegraphics[width=1.1\linewidth]{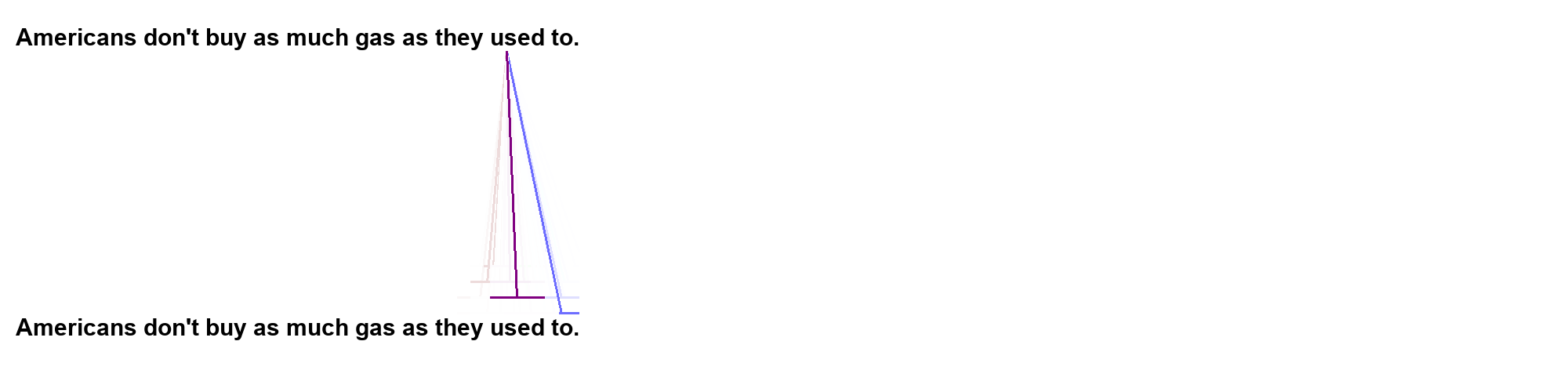}
      \caption{}
      \end{subfigure}
            \begin{subfigure}{\textwidth}
      \centering
      \includegraphics[width=1.1\linewidth]{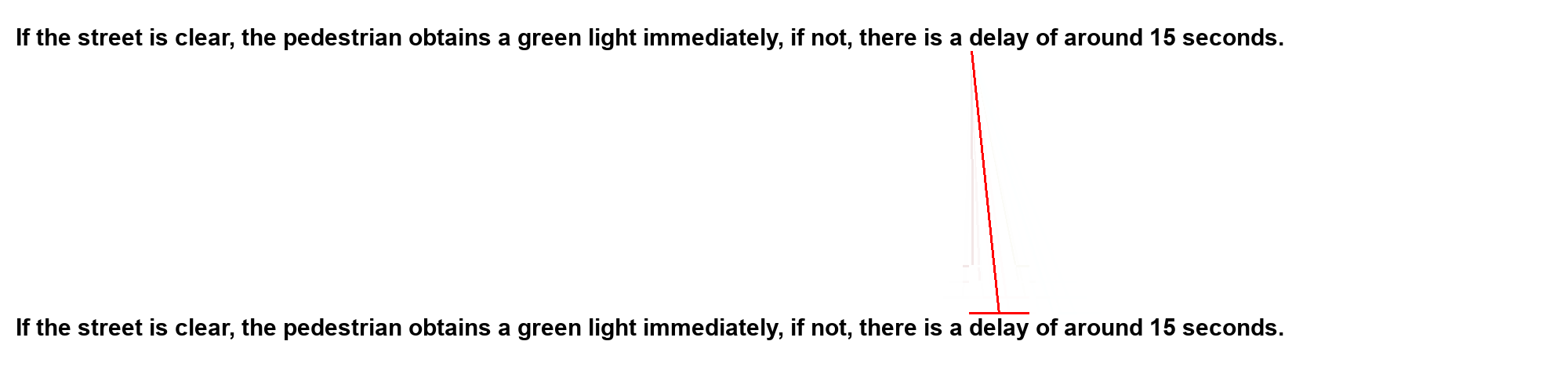}
      \caption{}
      \end{subfigure}
    \label{char}
    \caption{Examples of self-attention (using area attention) for the first layer of a 6-layer Transformer encoder, during character-level EN-DE translation tasks. For each example in Figure 2, the top row shows the query character and the bottom row shows the attended range of characters. The color indicates one of the 8 heads for the multi-head attention while the intensity shows the probability of the attention.}
    \end{adjustbox}
\end{figure}